\pdfoutput=1
\documentclass[11pt]{article}

\usepackage[]{emnlp2021}

\usepackage{times}
\usepackage{latexsym}

\usepackage[T1]{fontenc}
\usepackage[utf8]{inputenc}

\usepackage{microtype}

\usepackage{amssymb} % for \smallsetminus
\usepackage{graphicx}
\usepackage{capt-of}% or \usepackage{caption}
\usepackage{booktabs}
\usepackage{varwidth}
\usepackage{multirow, booktabs}
\newsavebox\tmpbox
\usepackage{mathtools}
\usepackage[linesnumbered,ruled,vlined]{algorithm2e}
\usepackage{algorithmic}

\SetKwInput{KwInput}{Input}                % Set the Input
\SetKwInput{KwOutput}{Output} 
\usepackage[inline]{enumitem}
\usepackage{amsmath}% http://ctan.org/pkg/amsmath

\title{Progressive Transformer-Based Generation of Radiology Reports}
\author{Farhad Nooralahzadeh\textsuperscript{1}, Nicolas Perez Gonzalez\textsuperscript{1}, Thomas Frauenfelder\textsuperscript{1},\\ \textbf{Koji Fujimoto\textsuperscript{2\textdagger}}, \textbf{Michael Krauthammer\textsuperscript{1}} \\
  \textsuperscript{1}University of Z\"urich and University Hospital of Z\"urich , \textsuperscript{2}Kyoto University \\
  \texttt{\{farhad.nooralahzadeh,nicolas.perez,michael.krauthammer\}@uzh.ch} \\  
  \texttt{thomas.frauenfelder@usz.ch,\textsuperscript{\textdagger}kfb@kuhp.kyoto-u.ac.jp}
  }

\date{}
\begin{document}
\maketitle

\begin{abstract}
Inspired by Curriculum Learning, we propose a consecutive (\emph{i.e.}, image-to-text-to-text) generation framework where we divide the problem of  radiology report generation into two steps. Contrary to generating the full radiology report from the image at once, the model generates global concepts from the image in the first step and then reforms them into finer and coherent texts using a transformer architecture. We follow the transformer-based sequence-to-sequence paradigm at each step. We improve  upon the state-of-the-art on two benchmark datasets.
\end{abstract}

\section{Introduction}

%The analysis of X-rays by radiologists in medical practice remains to be a challenging task. With years of training, these experts analyze images and recognize particular features that are later translated to a written report. This is a labor intensive and time consuming task, specially difficult for young trainees. With increasing demand, the burden on radiologists has increased over time, requiring the addition of the technologies to improve their workflow. Automatic report generation given an X-ray has significant potential to improve workflow in the clinical setting.

The analysis of X-rays in medical practice is the most common and important task for radiologists. With years of training, these experts learn to recognize particular features in the image that are later translated to a written report in a clinically appropriate manner. This is a labor intensive and time consuming task, especially difficult for young trainees. With increasing demand on imaging examinations, the burden on radiologists has increased over time, requiring the addition of the technologies to improve their workflow. 

Previous research on radiology report generation has mostly focused on image-to-text generation tasks. \citet{jing-etal-2018-automatic} introduced a co-attention mechanism to generate full paragraphs. \citet{lovelace-mortazavi-2020-learning} explored report generation through transformers. More recently, \citet{Zhang_2020} used a preconstructed graph embedding module on multiple disease findings to assist the generation of reports. Finally, \citet{chen-etal-2020-generating-radiology} proposed to generate radiology reports via memory-driven transformer and showed that their proposed approach outperforms previous models with respect to both language generation metrics and clinical evaluation. These systems have significant potential in many clinical settings, including improvement in workflow in radiology, clinical decision support, and large-scale screening using X-ray images.

%It is in this context that automated report generation has gathered attention in the recent years. Report generation in radiology aims to automatically generate descriptive text for given radiographic images. The primary objectives are to homogeneize clinical care, make the analysis of images reproducible and alleviate the burden on radiolgists. 

%Automatic report generation can be compared to other similar tasks such as image captioning but there are large differences between these problems. First, in a typical image captioning task such as COCO \citep{chen2015microsoft}, the captioning consists of short sentences whereas in report generation, the caption consists of multiple sentences composing a rather extensive paragraph. This sole point deems traditional captioning approaches \citep{vinyals2015show, anderson2018bottom} obsolete. In report generation the accuracy of positive or negative disease mentions is extremely important whereas in a more simple image captioning this might not be the case. Additionally, the mention of specific words is not important but rather the clinical relevance of a statement, which is lost with traditional metrics.

In this work, 
%we built on top of the first transformer model introduced by \citep{NIPS2017_3f5ee243}. 
we focus on generating reports from chest X-ray images innovating with a double staged transformer based architecture. %As stated by \citep{liu2019clinically}, most efforts in the past have guaranteed that the report looks real rather than predict accurately presence or absence of a disease. %The contribution from our work is the use of a two staged transformer based system, first identifying accurate labels and later enriching those initial sequences to more fluent sentences.
Our contributions in this paper can be summarized as follows: 
\begin{enumerate*}[label=(\roman*), itemjoin={{, }}, itemjoin*={{, and }}]
    \item We propose to produce radiology reports via a simple but effective progressive text generation model by incorporating high-level concepts into the generation process \footnote{Our code is available at \url{https://github.com/uzh-dqbm-cmi/ARGON}}
    \item We conduct extensive experiments and the results show that our proposed models outperforms the baselines and existing models, \emph{i.e.}, achieving a substantial $+1.23$\% increase in average over all language generation metrics in {\small \textsc{IU X-RAY}}, and the increase of $+3.2$\% F1 score in {\small \textsc{MIMIC-CXR}}, against the best baseline {\small \textsc{R2GEN}}
    \item We perform a qualitative analysis to further demonstrate the quality and properties of the generated reports.
\end{enumerate*}

\section{Method}
\begin{figure*}[t]
    \includegraphics[width=1.1\textwidth]{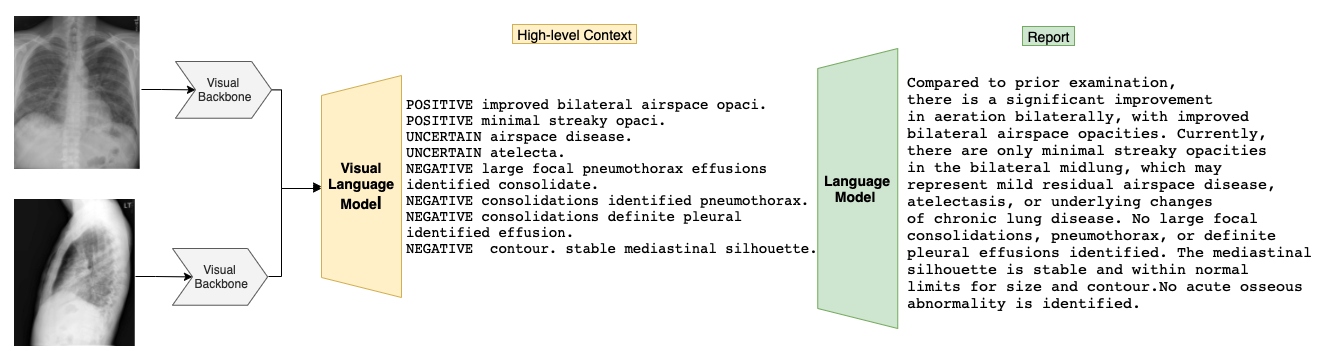}
    \caption{Overview of our proposed framework}
    \label{fig:framework}
\end{figure*}

An essential challenge in the radiology report generation is modeling the clinical coherence across the entire report.  Contrary to generating the full radiology report from the image at once, we propose a consecutive (\emph{i.e.}, image-to-text-to-text) generation framework (inspired by Curriculum Learning \citep{10.1145/1553374.1553380} and the work of \citet{tan2020progressive}). As shown in Figure \ref{fig:framework}, we divide the problem of  radiology report generation into two steps. In the first step, the model generates global concepts from the image and then reforms them into finer and coherent text using a transformer architecture. Each step follows the transformer based sequence-to-sequence paradigm. 
\paragraph{Model Architecture}
 Instead of generating the full report from an input radiology image, we frame the generation process such as: $X \rightarrow C \rightarrow Y$, where $X=\{x_1, x_2, ..., x_S\}$, $x_s \in \mathbb{R}^d$. $X$ is a radiology image  and $x_s$ is a sequence of patch features extracted from visual extractor and $d$ is the size of the feature vectors. $C=\{c_1, c_2, ..., c_T\}$, $c_t \in \mathbb{V}$, and $Y=\{y_1, y_2, ..., y_{T^{\prime}}\}$, $y_{t^{\prime}}  \in \mathbb{V^{\prime}}$, are the generated tokens at intermediate and final steps, respectively. $T$ and $T^{\prime}$ are the length of generated tokens and $\mathbb{V}$, $\mathbb{V}^{\prime}$ are the vocabulary of all possible tokens at each step. Our framework can be partitioned into three major components such as: 1) A visual backbone 2) An intermediate encoder-decoder as a visual language model (ViLM) and 3) A final encoder-decoder as a language model (LM).
\subparagraph{Visual Backbone} Given a set of radiology images ($I$), the visual backbone extracts the visual features $X$ and results in the source sequence $\{x_1, x_2, ..., x_s\}$ for the subsequent visual language model. The visual backbone can be formulated based on pre-trained Convolutional Neural Networks (CNN), e,g., DenseNet \citep{DBLP:journals/corr/HuangLW16a}, VGG \citep{DBLP:journals/corr/SimonyanZ14a} or ResNet \citep{DBLP:conf/cvpr/HeZRS16}. We find DenseNet to be more effective in our generation task and therefore use it as our based visual feature extractor. 
 \subparagraph{Visual Language Model (ViLM)}
 We adapt a state-of-the-art image captioning model, Meshed-Memory Transformer (\textsc{\small $\mathcal{M}^2$ Tr.}), introduced by \cite{cornia2020m2} for the intermediate step of our architecture. \textsc{\small $\mathcal{M}^2$ Tr.} is a transformer \citep{NIPS2017_3f5ee243} based model which presents two adjustments that leveraged the performance of the model: Memory Augmented Encoder and Meshed Decoder. Memory Augmented Encoder extends the set of keys and values in the encoder with additional “slots” to extract a priori information. The priori information is not based on the input; it is encoded in  learnable vectors, which are concatenated to keys and values and can be directly updated via SGD. Unlike the original decoder block in transformer, which only performs a cross-attention between the last encoding layer and the decoding layers, the \textsc{\small $\mathcal{M}^2$ Tr.} presents a meshed connection with all encoding layers. We refer the reader to \citet{cornia2020m2} for a detailed description of the Meshed-Memory Transformer.

Given the visual language model structure, the objective of the intermediate generation phase can be formalized as :
\setlength{\belowdisplayskip}{0pt} \setlength{\belowdisplayshortskip}{0pt}
\setlength{\abovedisplayskip}{0pt} \setlength{\abovedisplayshortskip}{0pt}
\[
p_{\theta}(C \mid I)=\prod_{t=1}^T p_{\theta}(c_t \mid c_{<t}, I)
\]
where $C$ at the intermediate step is the high-level context that contains informative and important tokens to serve as skeletons for the following enrichment process. 
To train the ViLM , we maximize the conditional log-likelihood $\sum_{t=1}^{T} \log p_{\theta}(C\mid I)$ on the training data to find the optimized $\theta^{*}$.
\subparagraph{Language Model}
The third component of our architecture is also based on the transformer as a sequence-to-sequence model that follows the conditional probability as: \[p_{\theta^{\prime}}(Y|C)=\prod_{t^{\prime}}^{T^{\prime}}p_{\theta^{\prime}} (y_{t^{\prime}}|y_{<t^{\prime}} \mid f_{\theta^{\prime}}(C))\]
where $f_{\theta}$ is an encoder that transforms the input sequence (\emph{e.g.}, high-level context) into another representation that are used by the language model $p_{\theta}$ at decoding step. We employed BART \citep{lewis-etal-2020-bart} as a pre-trained language model and fine-tune on our target domain. BART includes a BERT-like encoder and GPT2-like decoder. It has an autoregressive decoder and can be
directly fine tuned for sequence generation tasks such as paraphrasing and summarization. Similar to the previous module, to train the LM, we maximize the conditional likelihood $\sum_{t^{\prime}}^{T^{\prime}} \log p_{\theta^{\prime}}(Y\mid C)$ using the training set.

\paragraph{Training}
\begin{algorithm}[t]

\DontPrintSemicolon
 \KwInput{Radiology Reports \emph{R} and Images \emph{I}, Pretrained CNNs Model \emph{DensNet-121}, Pretrained LM \emph{BART}}
 \vspace{5pt}
 Extract a high-level context \emph{C} from Radiology Reports \emph{R} \\
  \vspace{5pt}
 
Fine-tune ViLM and LM independently\\
 \vspace{5pt}
\KwOutput{Fine-tuned ViLM and LM for report generation from Images \emph{I} in a progressive manner}
\caption{Training the Progressive Transformer-Based Generation of Radiology Reports}
\label{alg:1}
\end{algorithm}

Algorithm \ref{alg:1} shows the training steps of our proposed architecture.
We first extract a high-level context $C$ for each report in training dataset (see Figure \ref{fig:framework}). To do so, we employed \texttt{MIRQI} tools implemented by \citet{Zhang_2020}. Each training report is processed with disease word extraction, negation/uncertainty extraction, and attributes extraction based on dependency graph
parsing. A similar method proposed in NegBio \citep{Peng2018NegBioAH} and CheXpert \citep{irvin2019chexpert} for entity extraction and rule based negation detection is adopted in \texttt{MIRQI}. Then, we construct independent training data for each stage, \emph{i.e.}, fine-tuning of the ViLM and LM. More concretely, given training pairs $(I,C)$, we fine-tune ViLM. On the other hand, the BART is fine-tuned by using training pairs $(C, R)$ in the LM stage. 
Having fine-tuned the ViLM and LM, the model first generates the intermediate context and subsequently generates the full radiology report by adding finer-grained details at the final stage.
\section{Experiments}
\paragraph{Datasets}
We examine our proposed framework on two datasets as follows:
i) {\small \textsc{IU X-ray}} \citep{pub.1059743951}, a public radiology dataset that contains 7,470 chest X-ray images and 3,955 radiology reports, each report is associated with one frontal view chest X-ray image and optionally one lateral view image, ii) {\small \textsc{MIMIC-CXR}} \citep{DBLP:journals/corr/abs-1901-07042}, a large publicly available database of labeled chest radiographs that contains 473,057 chest X-ray images and 206,563 reports.
In order to compare our method with previous works, we use the available split on two datasets (\emph{i.e.}, the {\small \textsc{ IU X-Ray}} and {\small \textsc{MIMIC-CXR}} splits available in \citet{chen-etal-2020-generating-radiology}.)\footnote{\url{https://github.com/cuhksz-nlp/R2Gen}}
\paragraph{Evaluation Metrics}
The evaluation of the models is preformed using general NLG metrics including BLUE \citep{papineni-etal-2002-bleu}, METEOR \citep{denkowski-lavie-2011-meteor} and ROUGE-L \citep{lin-2004-rouge}. However, to address the shortcoming of the conventional NLG metrics in medical abnormality detection \citep{liu2019clinically,lovelace-mortazavi-2020-learning,chen-etal-2020-generating-radiology}, we also report clinical efficacy (CE) metrics that compare CheXpert extracted labels for the generated and reference reports\footnote{\url{https://github.com/MIT-LCP/mimic-cxr/tree/master/txt}}.      
To alleviate randomness of the scores, the mean of five different runs are reported.
\begin{table*}[t]
\centering
\setlength{\tabcolsep}{1.33mm}{\begin{tabular}{@{}l|l|cccccc|ccc@{}}
\toprule
\multirow{2}{*}{\textsc{\textbf{Data}}}                                             & \multicolumn{1}{c|}{\multirow{2}{*}{\textsc{\textbf{Model}}}} & \multicolumn{6}{c|}{\textsc{\textbf{NLG Metrics}}}                & \multicolumn{3}{c}{\textsc{\textbf{CE Metrics}}} \\
& \multicolumn{1}{c|}{}                                   & \textsc{BL-1}  & \textsc{BL-2}  & \textsc{BL-3}  & \textsc{BL-4}  & \textsc{MTR}   & \textsc{RG-L}  & \textsc{P}        & \textsc{R}       & \textsc{F1}      \\ \midrule
\multirow{2}{*}{\begin{tabular}[c]{@{}l@{}}\textsc{IU}\\ \textsc{X-Ray}\end{tabular}} & \textsc{Transformer}                           				                   &  0.388         &  0.246         &  0.176         &  0.133	       &    0.163      &        0.340           & -                 & -                & -                \\

& \textsc{$\mathcal{M}^2$ Tr.}                                             &    0.475      &     0.301      &        0.228   &      0.180    &     0.169      &   0.373                 & -                 & -                & -                \\
& \textsc{$\mathcal{M}^2$ Tr. Progressive}
&  \textbf{0.486} & \textbf{0.317} & \textbf{0.232} & \textbf{0.173} & \textbf{0.192}          & \textbf{0.390}    & -                 & -                & -                  \\ \midrule
\multirow{2}{*}{\begin{tabular}[c]{@{}l@{}}\textsc{MIMIC}\\ \textsc{-CXR}\end{tabular}} & \textsc{Transformer}                                              &    0.305        &  0.188          &      0.126     &      0.092    &     0.128     &  0.264                    & 0.313                & 0.224                & 0.261               \\

& \textsc{$\mathcal{M}^2$ Tr.} 
                                              &   0.361       &  0.221    &  0.146     &    0.101         &    0.139        &    0.266        &                 \textbf{0.324} & 0.241 &	 0.276     \\
 & \textsc{$\mathcal{M}^2$ Tr. Progressive}                         &                      \textbf{0.378} & \textbf{0.232} & \textbf{0.154} & \textbf{0.107} & \textbf{0.145} & \textbf{0.272}         & {0.240}& \textbf{0.428} &  \textbf{0.308}            \\ \bottomrule
\end{tabular}}
\vskip -0.3em
\caption{The performance of baseline and our progressive model on the test sets of \textsc{IU X-Ray} and {\small \textsc{MIMIC-CXR}} datasets with respect to NLG and CE metrics.
BL-n denotes BLEU score using up to n-grams; MTR and RG-L denote METEOR and ROUGE-L, respectively.
The performance of all models is averaged from five runs.}
\label{table:baselines}
\end{table*}

% ****************** Table 2 ******************
\addtolength{\tabcolsep}{-3pt}
\begin{table*}[t]
\centering
\resizebox{1\linewidth}{!}{
\begin{tabular}{@{}l|l|cccccc|ccc@{}}
\toprule
\multirow{2}{*}{\textsc{\textbf{Data}}}                                             & \multicolumn{1}{c|}{\multirow{2}{*}{\textsc{\textbf{Model}}}} & \multicolumn{6}{c|}{\textsc{\textbf{NLG Metrics}}}                                     & \multicolumn{3}{c}{\textsc{\textbf{CE Metrics}}} \\
                                                                                       & \multicolumn{1}{c|}{}                                         & \textsc{BL-1}  & \textsc{BL-2}  & \textsc{BL-3}  & \textsc{BL-4}  & \textsc{MTR}   & \textsc{RG-L}  & \textsc{P}     & \textsc{R}    & \textsc{F1}      \\ \midrule
\multirow{7}{*}{\begin{tabular}[c]{@{}l@{}}\textsc{IU}\\ \textsc{X-Ray}\end{tabular}}  & \textsc{ST}$^{\odot}$                                      & 0.216          & 0.124          & 0.087          & 0.066          & -              & 0.306          & -                 & -                & -                \\
                                                                                       & \textsc{Att2in}$^{\odot}$                                  & 0.224          & 0.129          & 0.089          & 0.068          & -              & 0.308          & -                 & -                & -                \\
                                                                                       & \textsc{AdaAtt}$^{\odot}$                                  & 0.220          & 0.127          & 0.089          & 0.068          & -              & 0.308          & -                 & -                & -                \\ 
                                                                                       & \textsc{CoAtt}$^{\odot}$                                   & 0.455          & 0.288          & 0.205          & 0.154          & -              & 0.369          & -                 & -                & -                \\
                                                                                       & \textsc{Hrgr}$^{\odot}$                                    & 0.438          & 0.298          & 0.208          & 0.151          & -              & 0.322          & -                 & -                & -                \\
                                                                                       & \textsc{Cmas-RL}$^{\odot}$                                 & 0.464          & 0.301          & 0.210          & 0.154          & -              & 0.362          & -                 & -                & -                \\ 
                                                                                        & \textsc{R2Gen}$^{\odot}$                                                & {0.470} & {0.304} & {0.219} & {0.165} & 0.187          & {0.371} & -                 & -                & -                \\  \cmidrule(l){2-11}
  %& \textsc{R2Gen}$^{\dagger}$ \small{(Avg of 5 runs)} & {0.431} & {0.267} & {0.184} & {0.135} & -          & {0.348} & -                 & -                & - \\

  &  \textsc{$\mathcal{M}^2$ Tr. Progressive}                                                & \textbf{0.486} & \textbf{0.317} & \textbf{0.232} & \textbf{0.173} & \textbf{0.192}          & \textbf{0.390}     & -                 & -                & - \\ \midrule
 % & \textsc{Ours} \small{(Avg of 5 runs)}                                                & \textbf{0.477} & \textbf{0.304} & \textbf{0.213} & \textbf{0.156} & -          & \textbf{0.362} & -                 & -                & -                \\\midrule
\multirow{5}{*}{\begin{tabular}[c]{@{}l@{}}\textsc{MIMIC}\\ \textsc{-CXR}\end{tabular}} & \textsc{ST}$^{\oplus}$                                        & 0.299          & 0.184          & 0.121          & 0.084          & 0.124          & 0.263          & 0.249             & 0.203            & 0.204            \\
                                                                                       & \textsc{Att2in}$^{\oplus}$                                    & 0.325          & 0.203          & 0.136          & 0.096          & 0.134          & 0.276          & 0.322             & 0.239            & 0.249            \\
                                                                                       & \textsc{AdaAtt}$^{\oplus}$                                    & 0.299          & 0.185          & 0.124          & 0.088          & 0.118          & 0.266          & 0.268             & 0.186            & 0.181            \\
                                                                                       & \textsc{Topdown}$^{\oplus}$                                   & 0.317          & 0.195          & 0.130          & 0.092          & 0.128          & 0.267          & 0.320             & 0.231            & 0.238            \\ 
 
 & \textsc{R2Gen} $^{\odot}$
 & {0.353} & {0.218} &  {0.145} &  {0.103} &  {0.142} &  \textbf{0.277}  & \textbf{0.333}    & {0.273}   & {0.276}   \\ \cmidrule(l){2-11} 
 &  \textsc{$\mathcal{M}^2$ Tr. Progressive} 
 & \textbf{0.378} & \textbf{0.232} & \textbf{0.154} & \textbf{0.107} & \textbf{0.145} & {0.272} &  {0.240} & \textbf{0.428}&   \textbf{0.308}         \\\bottomrule
\end{tabular}
}
%\vskip -0.3em
\caption{Comparisons of our full model with previous studies on the test sets of \textsc{IU X-Ray} and {\small \textsc{MIMIC-CXR}} with respect to language generation (NLG) and clinical efficacy (CE) metrics.
$\odot$ refers to the result that is directly cited from the original paper and $\oplus$ represents the replicated results reported on \citet{chen-etal-2020-generating-radiology}.}
\label{table:2}
%\vskip -1em
\end{table*}
\paragraph{Baselines}
We consider the following baselines in our evaluation process:
\begin{enumerate*}[label=(\roman*), itemjoin={{, }}, itemjoin*={{, and }}]
    \item \textbf{{\small\textsc{Transformer}}}: The vanilla transformer is employed in the ViLM component to generate radiology reports in a standard manner
    \item \textbf{{\small\textsc{$\mathcal{M}^2$ Tr.}}}: The Meshed-Memory Transformer is used in the ViLM component to generate text without progressive style.  
\end{enumerate*} 

Moreover, we compare our model with previous studies reported
in \citet{chen-etal-2020-generating-radiology}, \emph{e.g.}, \textbf{{\small\textsc{ST}}} \cite{vinyals2015show}, \textbf{{\small\textsc{Att2in}}} \cite{scst}, \textbf{{\small\textsc{AdaAtt}}} \cite{adaatt},  \textbf{{\small\textsc{Topdown}}} \cite{anderson2018bottom}, \textbf{{\small\textsc{CoATT}}} \cite{jing-etal-2018-automatic}, \textbf{{\small\textsc{Hrgr}}} \cite{hrgr}, \textbf{{\small\textsc{Cmas-Rl}}} \cite{cmas} and \textbf{{\small\textsc{R2GEN}}} \cite{chen-etal-2020-generating-radiology} (see Section \ref{previous} in appendix for more detail).
For reproducibility, the model configuration and training are described in Section \ref{implementation} of the Appendix.
\section{Results and Discussion}
\paragraph{Effect of progressive generation} To show the effectiveness of our model, we conduct experiments with baseline models, including our proposed model (\emph{i.e.}, {\small \textsc{$\mathcal{M}^2$ Tr. Progressive}} ) as reported in Table \ref{table:baselines}.
The results shows that {\small \textsc{$\mathcal{M}^2$ Tr.}} provides better performance than the vanilla transformer which confirms the validity of incorporating memory matrices in the encoder and meshed connectivity between encoding and decoding modules. Our progressive model consistently outperforms the standard and single-stage ViLMs by a large margin on almost all metrics in both benchmark datasets, which clearly highlights the benefits of the progressive generation strategy. However the precision of the progressive model is lower than the baselines. We observe that the progressive generation produces long reports mostly by adding the abnormality mentions in negation mode (\emph{e.g.}, \emph{No evidence of pneumonia}, \emph{There is no pneumothorax}
), therefore it increases the number of false positives (FPs) in the CE metrics.

In Table \ref{table:2}, we compare our full model (\emph{i.e.},  {\small \textsc{$\mathcal{M}^2$ Tr. Progressive}}) with the previous works on the same datasets.
In general, memory based transformer methods offer significant improvements across all metrics compared to the recurrent neural networks (RNNs) based architectures. This is illustrated by comparing {\small \textsc{R2GEN}}, {\small \textsc{$\mathcal{M}^2$ Tr.}} and our full model with the other techniques (see also Table \ref{table:baselines}). Our model achieves competitive results compare to  {\small\textsc{R2GEN}}, \emph{i.e.}, $+1.23\%$ average on all NLG metrics in {\small \textsc{IU X-ray}}, $+0.83\%$ and $+3.2\%$ average on all NLG metrics and F1 score, respectively, in the {\small \textsc{MIMIC-CXR}} dataset.
This indicates the benefits of using the {\small \textsc{$\mathcal{M}^2$ Tr.}} together with our progressive strategy in the radiology reports generation task. We hypothesise that the use of \texttt{MIRQI} in the intermediate context generation provides informative and high-quality plans which results in reasonable descriptions for clinical abnormalities in the last generation stage. 
\paragraph{Analysis}
As a qualitative analysis to explain the effectiveness of our progressive model, we examine some of the generated reports with their references from the {\small \textsc{MIMIC-CXR}} test dataset (see Figure \ref{fig:2} in the Appendix). We show the text alignments between the reference text and generated one with the same colors.
It can be seen in the top two examples the progressive model is able to provide reports aligned with the reference texts where the baseline model fails to cover them, \emph{e.g.}, \emph{post median sternotomy}, and \emph{mitral valve replacement}, \emph{The mediastinal contours}, \emph{ enlargement of the cardiac silhouette}, \emph{ bilateral pleural effusions} and \emph{compressive atelectasis} in the top two examples are not generated by {\small \textsc{$\mathcal{M}^2$ Tr.}}.  Although our model shows improvements in the NLG and CE metrics evaluation, it still fails to generate clinically coherent and error-free reports. For example, in the third example of Figure \ref{fig:2}, the \emph{mild pulmonary edema} is incorrect since the \emph{No new parenchymal opacities} in the reference implies negative pulmonary edema. Furthermore, the sentence \emph{left plueral effusion} in the last example is not consistent with the previous text \emph{bilateral pleural effusion}.  
Additionally, the examples in Figure \ref{fig:2} contain a comparison of study against to the previous study such as \emph{ As compared to the previous ...} and \emph{ In comparison with the study ...} in the generated reports. This is a little surprising since the model does not have any clue about the previous report of a patient in its design. It can be attributed to the fact that these template sentences are more frequent in the training set. 
The examples also show that the progressive model generates a more comprehensive report compare to the baseline.It includes occasionally the extra mentions of medical terms compared to the reference text (\emph{e.g.}, \emph{ There is no focal consolidation} and \emph{ No evidence of pneumonia} in examples 1 and 3, respectively), which result in false-positive mention of observations in the CheXpert labeler of the CE metrics.

\section{Conclusion}
We propose to produce radiology report via a simple but effective progressive text generation model by incorporating high-level concepts into the generation process. The experimental results show that our proposed model outperforms the baselines and a wide range of radiology report generation methods, in terms of language generation and clinical efficacy metrics. Further, the manual analysis demonstrates the ability of the model to produce long and more clinically coherent reports, however there is still room for improvement.   
\section*{Acknowledgements}
This research has received funding from the Kyoto University-University of Zurich Joint Funding Program. The authors would also like to thank the anonymous reviewers for their feedback.
\bibliography{emnlp2021}
\bibliographystyle{acl_natbib}
%++++++++++++++++++++++++++++++++
\clearpage
\appendix

%******************** Figure 2 ********************
\begin{figure*}[t]
\centering
\includegraphics[width=1\textwidth, trim=0 20 0 0]{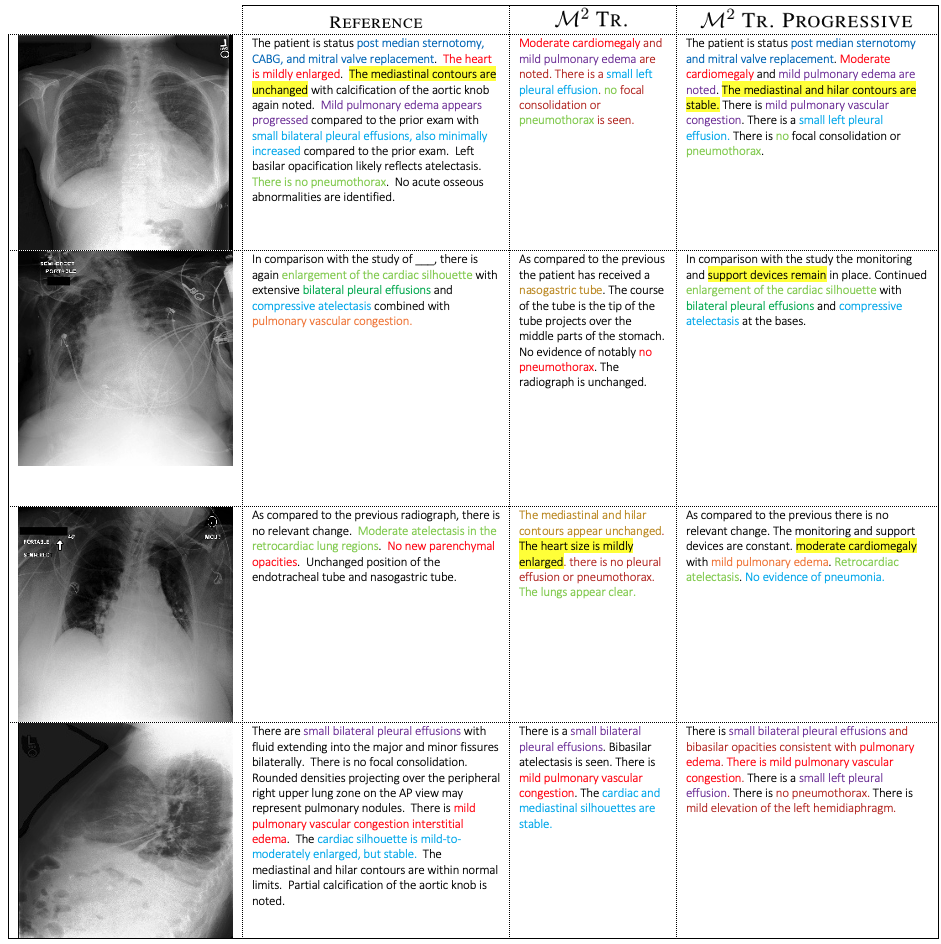}
\caption{Illustrations of reports from test dataset as {\small \textsc{Reference}}, {\small \textsc{$\mathcal{M}^2$ Tr.}} as a baseline model  and {\small \textsc{$\mathcal{M}^2$ Tr. Progressive}} as a proposed model for selected X-ray chest images. Different colors highlight different medical terms and the detected abnormalities. The text alignments between the reference text and generated one are highlighted with the same colors. Top two images are positive results, the bottom two ones are partial failure cases.}
\label{fig:2}

\end{figure*}
\section{Previous Models} \label{previous}
\begin{itemize}
    \item \textbf{\textsc{ST}} \cite{vinyals2015show}:  The model is based on a convolution neural network that encodes an image into a compact representation, followed by a recurrent neural network that generates a corresponding sentence. The model is trained to maximize the likelihood of the sentence given the image.
    \item \textbf{\textsc{Att2in}} \cite{scst}: The CNN-RNN based model which rather than utilizing a static, spatially pooled representation of the image, it employs the attention model. The attention model dynamically re-weight the input spatial (CNN) features to focus on specific regions of the image at each time step. The model considers a modification of the architecture of the attention model for captioning in \citet{Xu2015ShowAA}, and input the attention-derived image feature only to the cell node of the LSTM.
    \item \textbf{\textsc{AdaAtt}} \cite{adaatt}: It is an adaptive attention encoder-decoder framework which provides a fallback option to the decoder. At each time step, the model decides whether to attend to the image (and if so, to which regions) or to the visual sentinel. The model decides whether to attend to the image and where, in order to extract meaningful information for sequential word generation.
    \item \textbf{\textsc{Topdown}} \cite{anderson2018bottom}: A combined bottom-up and top-down visual attention mechanism (based on Faster R-CNN). The bottom-up mechanism proposes image regions, each with an associated feature vector, while the top-down mechanism determines feature weightings. The model enables attention to be calculated more naturally at the level of objects and other salient regions.
    \item \textbf{\textsc{CoATT}}
    \cite{jing-etal-2018-automatic}: A multi-task learning framework which jointly performs the prediction of tags and the generation of paragraphs. The model is based on a hierarchical LSTM model and incorporates a co-attention mechanism to localize regions containing abnormalities and generate narrations for them.
    \item \textbf{\textsc{Hrgr}} \cite{hrgr}: A Hybrid Retrieval-Generation Reinforced Agent consists of a CNN to extract visual
    features which is then transformed into a context vector by an image encoders. Then a sentence decoder (RNNs-based with attention mechanism) recurrently generates a sequence of hidden states which represent sentence topics. A retrieval policy module is employed to decide for each topic state to either automatic generate a sentence, or retrieve a specific template from a template database. 
    \item \textbf{\textsc{Cmas-Rl}} \cite{cmas}:
    It is a LSTM based framework for generating chest X-ray imaging reports by exploiting the structure information in the reports. It explicitly models the between-section structure by a two-stage framework, and implicitly captured the within-section structure with a Co-operative Multi-Agent System (CMAS) comprising three agents: Planner (PL), Abnormality Writer (AW) and Normality Writer (NW). The entire system was trained with REINFORCE algorithm.
    \item \textbf{\textsc{R2GEN}} \cite{chen-etal-2020-generating-radiology}: The model uses ResNet as a visual backbone and generate radiology reports with memory-driven Transformer, where a relational memory is designed to record key information of the generation process and a memory-driven conditional layer normalization is applied to incorporating the memory into the decoder of Transformer. It obtained the state-of-the-art on two radiology report datasets.  
\end{itemize}
\section{Implementation detail} \label{implementation}
We adopt the codebase of {\small \textsc{R2Gen}}\footnote{\url{https://github.com/cuhksz-nlp/R2Gen}} to implement our proposed model.
We use DenseNet121 \cite{DBLP:journals/corr/HuangLW16a} 
pre-trained on CheXpert dataset with 14-class classification setting \footnote{Available in \url{https://nlp.stanford.edu/ysmiura/ifcc/chexpert_auc14.dict.gz}}, as the visual backbone to extract visual features with the dimension 1024.
For {\small \textsc{IU X-RAY}}, the two images are employed to guarantee fair comparison with previous works. In ViLM component, we use the {\small \textsc{$\mathcal{M}^2$ Tr.}} \cite{cornia2020m2} with $8$ attention head, memory size equal to $40$, and $3$ encoder layers and decoder layers. The model dimension is $512$ with the feed forward layers have a dimension of $2048$. In LM component, we adapt a pre-trained
BART, \emph{i.e.}, \texttt{bart-base}\footnote{Available in \url{https://huggingface.co/facebook/bart-base}} for generation of final reports. 
The model is trained with the Adam optimiser with batch size of $16$. The learning rates are set to $5e-5$ and $1e-4$ for the visual extractor and the remaining parameters, respectively. The maximum length in {\small \textsc{IU X-RAY}} is set to $60$ and  in {\small \textsc{MIMIC-CXR}} is set to $100$. Beam search with beam size of $3$ and $5$ is used to decode texts during experiments with {\small \textsc{IU X-RAY}} and  {\small \textsc{MIMIC-CXR}}, respectively.
The hyper-parameters values are obtained by evaluation of the model with the best \texttt{BLEU-4} score using the validation set of two benchmark datasets. 
We train the model using NVIDIA GeForce RTX 2080 Ti for 100 and 30 epochs with early stopping (patience=20) on {\small \textsc{IU X-RAY}} and {\small \textsc{MIMIC-CXR}}, respectively.

\end{document}